\documentclass[preprint,12pt,authoryear]{elsarticle} 

\usepackage{graphicx} 
\usepackage[graphicx]{realboxes}
\usepackage{bm} 

\usepackage{subcaption}
\usepackage{lipsum}
\usepackage[colorlinks]{hyperref}
\usepackage{xcolor} 
\usepackage{multirow}
\usepackage{array}
\usepackage{gensymb} 
\usepackage{lineno}
\modulolinenumbers[5]

\usepackage{lscape}

\makeatletter
\providecommand{\doi}[1]{%
  \begingroup
    \let\bibinfo\@secondoftwo
    \urlstyle{rm}%
    \href{http://dx.doi.org/#1}{%
      doi:\discretionary{}{}{}%
      \nolinkurl{#1}%
    }%
  \endgroup
}
\makeatother

\makeatletter
\newif\ifhbonecolumn
\@ifclasswith{emulateapj}{onecolumn}{\hbonecolumntrue}{\hbonecolumnfalse}
\makeatother

\newcolumntype{C}[1]{%
>{\centering}m{#1}}%

\begin{document}

\begin{frontmatter}

\title{\LARGE \bf
The Time-SIFT method : detecting 3-D changes from archival photogrammetric analysis with almost exclusively image information
}
\author[LISAH]{D.~Feurer\corref{cor1}}\ead{denis.feurer@ird.fr}
\author[LISAH]{F.~Vinatier}

\address[LISAH]{LISAH, Univ Montpellier, INRA, IRD, Montpellier SupAgro, Montpellier, France}
\cortext[cor1]{Corresponding author}

\begin{abstract}

Archival aerial imagery is a source of worldwide very high resolution data for documenting paste 3-D changes. However, external information is required so that accurate 3-D models can be computed from archival aerial imagery. In this research, we propose and test a new method, termed Time-SIFT (Scale Invariant Feature Transform), which allows for computing coherent multi-temporal Digital Elevation Models (DEMs) with almost exclusively image information. This method is based on the invariance properties of the SIFT-like methods which are at the root of the Structure from Motion (SfM) algorithms. On a test site of 170 km\textsuperscript{2}, we applied SfM algorithms to a unique image block with all the images of four different dates covering forty years. We compared this method to more classical methods based on the use of affordable additional data such as ground control points collected in recent orthophotos. We did extensive tests to determine which processing choices were most impacting on the final result. With these tests, we aimed at evaluating the potential of the proposed Time-SIFT method for the detection and mapping of 3-D changes. Our study showed that the Time-SIFT method was the prime criteria that allowed for computing informative DEMs of difference with almost exclusively image information and limited photogrammetric expertise and human intervention. Due to the fact that the proposed Time-SIFT method can be automatically applied with exclusively image information, our results pave the way to a systematic processing of the archival aerial imagery on very large spatio-temporal windows, and should hence greatly help the unlocking of archival aerial imagery for the documenting of past 3-D changes.
 \end{abstract}


\end{frontmatter}


\section{INTRODUCTION}

For decades, aerial imagery was used to produce large-scale geographic and topographic maps. With - most often - sub-metric resolutions, archival aerial imagery has hence a remarkable coverage both in the temporal and spatial dimensions. Archival aerial imagery exists in almost every country in the world from the first half of the twentieth century \citep{Cowley2012}. This imagery was most often acquired with stereoscopic coverage, which results in a unequalled potential for 3-D documentation of past changes. Besides, during the last decade 3-D past changes have been studied using archival aerial imagery in the disciplines of archaeology \citep{Verhoeven2011,Sevara2013,Verhoeven2016,Salach2017,Sevara2017}, geomorphology \citep{Gomez2015,Goncalves2016,Ishiguro2016,Bakker2017}, glaciology \citep{Mertes2017,Molg2017,Vargo2017}, and volcanology \citep{Gomez2014}.

However, the full potential of this huge volume of historical data may have not been widely exploited yet. This is mainly due to the fact that accurate processing of historical aerial imagery time series for 3-D change assessment requires other information than the images themselves. Indeed, external information such as camera calibration certificates, ground control points (GCPs), or the a priori knowledge of stable zones is needed to estimate exterior and interior orientation of the image blocks. This is all the more important when the aim is to compute Digital Elevation Models (DEMs) of differences (abbreviated into DoD by \cite{Lane_2003,Wheaton2010,Williams2012}), which require that differentiated DEMs are both accurate and spatially consistent between each other.\\ 

Historically, the construction of 3-D models began with photogrammetry, firstly with stereoscopic analysis of oriented image pairs and then with automatic correlation of oriented images \citep[see for example][]{Kraus1998}. Using these 'classical' photogrammetric methods, there are several examples of authors succeeding in differencing DEMs within most favourable conditions, i.e when image datasets are associated with all necessary calibration certificates \citep{Fabris2005,Fischer2011,Micheletti2015,Aucelli2016,Fieber2018}. Having calibration certificates is not sufficient, though. \cite{Fischer2011} and  \cite{Fieber2018} - with a specific case implying DoD with a satellite DEM for the latter work - needed to proceed additional co-registration of the photogrammetric DEMs, with a priori knowledge and manual delimitation of stable areas when necessary. Moreover, the use of 'classical' photogrammetric software, requires accurate initial information (interior  orientation, ground control points) so that processing succeeds. For instance, in a work realised with the ERDAS LPS software, \cite{Micheletti2015} had to give a specific care to ground control points choice, positioning, distribution and accuracy.
 
When calibration certificates are missing, which is not rare for oldest images, interior geometry can be estimated through autocalibration, which estimates simultaneously interior and exterior orientations. This method was successfully used by several authors with a view to compute DoDs thereafter \citep[e.g.][]{Chandler1989,Walstra2007,Redweik2016}. In their pioneering work, \cite{Chandler1989} reminded that interior orientation consists in two groups of parameters: the parameters characterising the displacement of the principal point relatively to the fiducial marks, and the parameters characterising the lens distortion. To these parameters has to be added the deformation due to the scanning, which has a strong impact on the performance of 3-D estimation \citep{Sevara2016}. These additional unknowns severely increase the need for ground control points, which in turn raises new issues : the need for expensive additional ground survey and the problem of the accuracy of orientation estimation with self-calibration algorithms \citep{Aguilar2013}.

A frequent strategy to collect these additional GCPs at limited cost is to rely on recent data. These are usually of better quality, associated with existing and accurate calibration information, fiducial marks and/or even contemporary GCPs. \cite{James2006} extracted different numbers of required GCPs from a shaded-view of a lidar DEM and then detected 3-D changes from photogrammetric DEMs estimated with older images. Several authors exploited a well-known or well-established geometry of recent image blocks, extracted GCPs from these recent images and used these GCPs in older imagery in order to determine interior and exterior orientation of these older image blocks \citep{Hapke2005,Dewitte2008,Zanutta2006,Fox2008}. More recently, \cite{Giordano2018} proposed a method allowing for an automatic collection of these GCPs. GCPs are detected in a recent orthophoto and DEM and then transferred to images of previous dates. The method relies on the detection of keypoints between images of different dates, which is made possible through a first estimation of coarse DEMs and orthophotos of ancient dates. These first estimates are done on the basis of available image metadata at all dates.

The advent of Structure from Motion (SfM) and Multi View Stero (MVS) algorithms in geoscience and archaeology in the last decade provided a new paradigm. Indeed, as explained for instance by \cite{Westoby2012} or \cite{Fonstad2013}, SfM algorithms compute relative orientations with image information exclusively, thanks to keypoints detection algorithms such as the Scale Invariant Feature Transform (SIFT) \citep{Lowe2004}. The potential of SfM-MVS algorithms for 3-D exploitation of archival aerial imagery was detected very soon by \cite{Verhoeven2011} who succeeded in obtaining a visual 3-D model from historical aerial imagery with exclusively image information. Moreover, the use of SfM software usually does not require advanced skills in photogrammetry, which allowed a wider use of these techniques, as noticed by recent reviews \citep{Eltner2016,Smith2016,Mosbrucker2017}.
However, even if SfM algorithms have allowed for a more accessible and more straightforward use of archival aerial imagery, it did not eliminate the need for thorough control of image acquisition geometry for differencing DEMs computed with such imagery. \cite{Bakker2017} has noted that SfM-MVS photogrammetric processing shares the limitations of classical methods : the propagation of linear errors into the final DEMs requires specific correction procedures for subsequent accurate DEM differencing. Once again, control of image and DEM geometry in SfM-based workflows was done through the use of additional external information. Several authors have used GCPs to help the autocalibration \citep{Gomez2014,Verhoeven2016,Mertes2017}. Other authors have co-registrated the DEMs obtained from archival aerial imagery onto reference DEMs, with an a priori knowledge of stable zones \citep{Bakker2017,Mertes2017,Molg2017,Sevara2017}.

Finally, as remarked for instance by \cite{James2017} in the context of unmanned aerial vehicles imagery processing, it is worth noting that quality and consistency of surveys realised with SfM-MVS algorithms still rely on an in-depth adjustment of processing parameters. Parameters adjustment may be even more complex and sensitive when dealing with interior orientation of archival aerial imagery due to the additional degrees of freedom of the problem. Thus, and even if SfM-MVS algorithms had made 3-D exploitation of aerial imagery more widely accessible, processing of archival imagery has still required manual intervention to avoid gross errors. This has been done by visual inspection throughout the whole processing \citep[e.g.][]{Bakker2017} or with complex strategies for the choice of parameters \citep[e.g.][]{Verhoeven2016}.

Thus, even if SfM photogrammetry should allow for a broader use of historical aerial imagery, all existing methods for extracting 3-D changes from this archive still rely on external data, significant expertise and/or parameters optimisation. This need for expertise and additional data still impedes the actual unlocking of big archival aerial imagery datasets. There is hence a need for a method that would allow the automated production of DoDs from archival aerial imagery with minimal external data and expertise.\\

In this context, our study aims at assessing a new method that takes benefit of the SIFT-like algorithms capabilities for a new purpose. Invariance of features detected by SIFT-like algorithms is originally spatial. It consists in invariance to scale and to simple geometric transforms. Our work lies on the fact that SIFT-like algorithms also show invariance in time, as noticed for instance by \cite{Chanut2017}, and by \cite{Vargo2017}. The principle of our proposed method, hence termed Time-SIFT, is to merge all dates of a collection of historical image archives in the same SfM processing. Our work proposes and tests an automatic method where images of different dates are processed all together in the same block during the SfM step. As a result, all dates share a same unique geometric reference, which saves the expensive and/or cumbersome collection of additional external data. Moreover, putting all images in the same block diminishes the linkage between interior end exterior orientation unknowns, hence allowing the use of SfM algorithm with no a priori photogrammetric expertise nor complex strategies for the choice of SfM parameters.

To the best of our knowledge, there is no existing study about the potential of a method which would systematically gather images of different dates in the first steps of SfM algorithms for DEM differencing. The aim of our paper is hence to assess the potential of the proposed Time-SIFT method to automatically obtain informative DoDs with limited input from the user, i.e with almost no other information that image information and with limited expertise in photogrammetric processing.

In this paper, we propose a test of the different classical processing methods usually used in the literature as compared with the proposed Time-SIFT method. 3-D models and DoDs are computed with different processing options and their quality is then assessed with independent DEMs. This analysis allows to determine the most significant processing options. A visual inspection of specific 3-D changes detected in the DoDs is also done in order to ascertain the capacity of the proposed method for mapping past 3-D changes.

\section{Material and methods}

\subsection{Test site and image data}

\begin{figure}
	\centering
	\includegraphics[width=\linewidth]{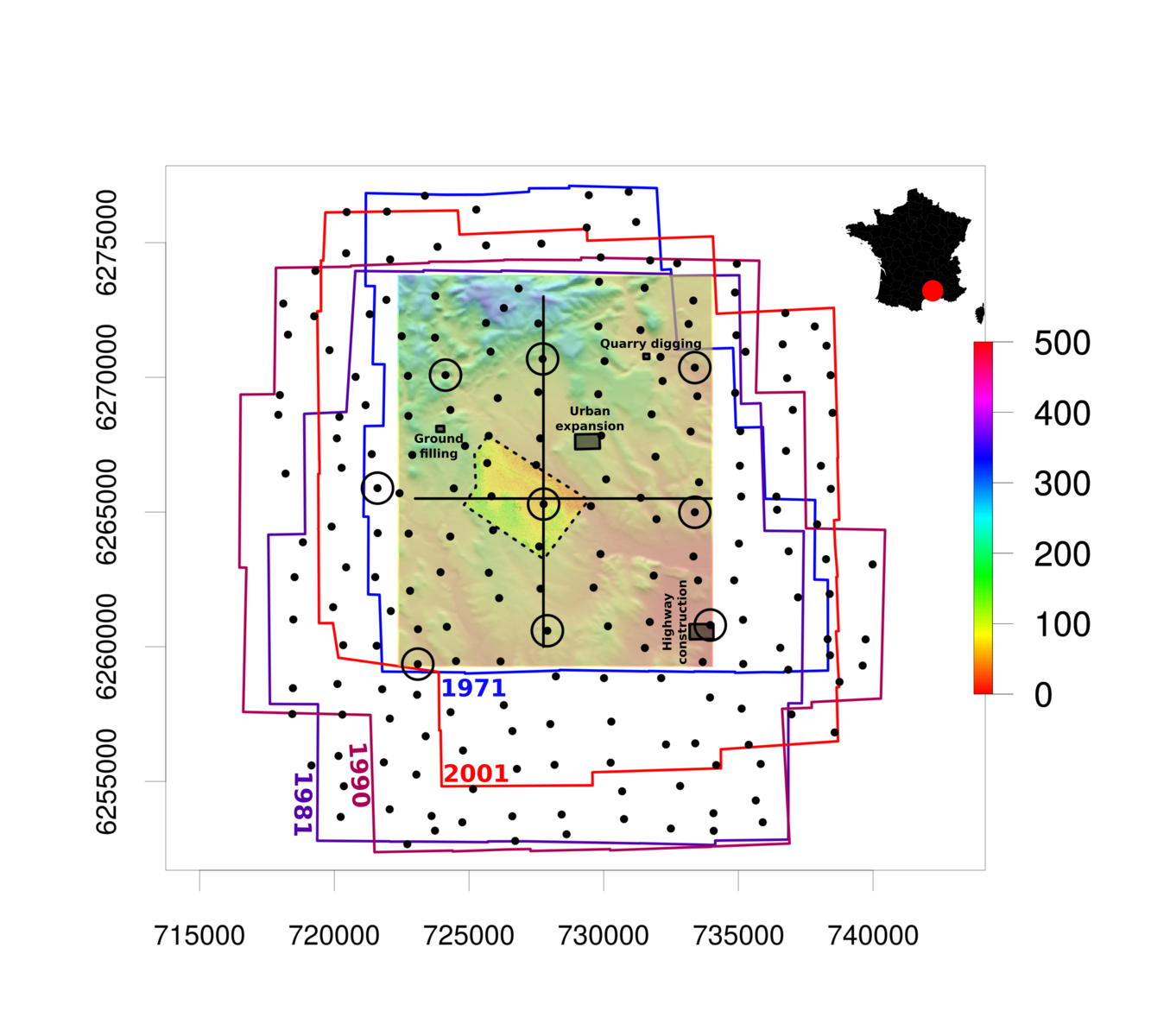}
	\caption{Localisation of the test area, image data, external DEMs and additional points. The coarse DEM used for validation, represented in colour scale in the middle of figure, covers the whole test area of 170 km\textsuperscript{2}. The smaller area bordered by dashed lines corresponds to the second validation DEM, represented with the same color scale. Black points represent the 200 manually marked points. Points belonging to the minimal set of 9 manually marked points are circled. Shaded rectangles highlight the zones where full resolution qualitative analysis of the DoDs was done. Two perpendicular lines correspond to the North-South and East-West transects. Approximate footprints of the image blocks are represented by coloured polygons. Map projection : Lambert 93 (EPSG:2154)}
  \label{fig:test_site_DEMs_GCPs_transects}
\end{figure}

The test site is a 170 km\textsuperscript{2} rectangle of a Mediterranean landscape located in Occitanie in southern France (43\degree 5N, 3\degree 19E). This zone is mainly covered by vineyards with forests in uppermost parts. This zone exhibited a change in land management through a severe transformation of vineyards areas during the 1980s, from goblet to trellised vineyards, with a progressive land abandonment and urbanization in the last 50 years \citep{Vinatier2018}. Altitudes vary between 0 and 350 m (Figure \ref{fig:test_site_DEMs_GCPs_transects}).
Archival aerial images were recently released by the IGN, the French national geographic institute and are freely available on the IGN's website (\url{https://remonterletemps.ign.fr/}). These images were scanned with photogrammetric scanners, which is important for a further photogrammetric processing as noted by \cite{Sevara2016} and also observed in a preliminary study \citep{Feurer2017SFPT}. We used a sample of these data at four different dates allowing for a stereo coverage of the entire test site (coloured polygons on Figure \ref{fig:test_site_DEMs_GCPs_transects}). Characteristics of these images are reported in Table \ref{tab:samples}.

\begin{table}
\centering
\scriptsize{
\begin{tabular}{l    C{.4cm} C{.9cm} C{.9cm} C{.5cm} C{.4cm} C{1cm}}

Date  & Focal length (mm)  & Estimated flight height (m) & Estimated scale & Images (\#) & Scan size ($\mu$m) & Estimated ground resolution (cm) \tabularnewline
\hline
21-06-1971 & 152 & 2700 & 1/18000 & 61 & 21 & 37 \tabularnewline
16-06-1981 & 153 & 4800 & 1/32000 & 27 & 21 & 66 \tabularnewline
25-06-1990 & 153 & 5000 & 1/32000 & 31 & 21 & 69 \tabularnewline
04-06-2001 & 153 & 4000 & 1/26000 & 44 & 21 & 55 \tabularnewline
\end{tabular}
}
\caption{Characteristics of image data.}
\label{tab:samples}
\end{table}

\subsection{DEMs and DoDs computing}\label{sec:DEM_DoD_computing}

DEM computing was done with AgiSoft Photoscan Pro\textsuperscript{\copyright} 1.2.6, which follows a SfM-MVS workflow. Processing parameters are reported in Table \ref{tab:photoscanfixed}. First, cameras were aligned. This step begins with an automated image features detection and matching which is comparable to the SIFT algorithm \citep{PhotoscanAlgo}. Based on the automatically matched points between images, image interior and relative exterior orientations were computed. Then an optimization step where autocalibration is refined was performed. Depending on the different processing options described in section \ref{sec:additionnal_points} below, additional manually marked GCPs and/or tie points could be used at this step. Once the image block geometry was refined, the dense image matching step was done and resulted in the estimation of a dense 3-D point cloud. Finally, DEMs were exported with a 1 m ground sampling distance with disabled interpolation to avoid subsequent processing artefacts in the DoDs. DoDs were finally computed by subtracting two successive DEMs. No-data values in at least one of the two input DEMs resulted in no-data values in the computed DoD.

\begin{table}
\centering
\scriptsize{
\begin{tabular}{l l l}
Processing step & Property & Value \\
\hline
Alignment & Accuracy & Highest\\
& Pair preselection & Disabled\\
& Key point limit & 100,000\\
& Tie point limit & 50,000\\
& Adaptive camera model fitting & Yes\\
\hline
Optimization & Lens parameters & f,b1,b2,cex,cy\\
& & k1,k2,k3,k4,p1,p2\\
& Marker accuracy (pix) & 5\\
& Marker accuracy (m) & 20\\
& Tie point accuracy (pix) & 1\\
\hline
Dense cloud & Quality & High\\
& Depth filtering & Moderate\\
\hline
Mesh & Surface type & Arbitrary\\
& Interpolation & Disabled\\
& Quality & Medium\\
& Depth filtering & Moderate\\
\hline
DEM & Pixel size (m) & 1\\
    & Interpolation & Disabled\\
\end{tabular}
}
\caption{Parameters used in PhotoScan Pro\textsuperscript{\copyright}.}
\label{tab:photoscanfixed}
\end{table}

\subsection{Investigated processing options}\label{sec:processing_options}

The interest of the proposed Time-SIFT method was evaluated by testing alternative processing options at various steps of the SfM-MVS processing chain. Apart from the two options of using or not the Time-SIFT method, three image pre-processing options, three options for the use of additional manually marked points and two autocalibration options were investigated. Details about these options is given in the following subsections. These different processing options resulted in 36 different processing scenarios and hence, with 4 dates, 144 DEMs and 108 DoDs.

\subsubsection{Image pre-processing}

Converting image coordinates into camera coordinates requires to consider (i) the film movement in the camera body between two image acquisitions (ii) possible film deformations during storage and (iii) possible geometric deformation during film scanning. Fiducial marks are usually used to estimate the geometric transform between image coordinates and camera coordinates. Another point, raised by \cite{Gomez2015} and questioned by \cite{Bakker2017}, is the fact that image borders, where fiducial marks and other metadata appear, can interfere with the automated feature detection algorithm used in SfM workflows and may need to be masked. 

Three options were hence tested. The first option, namely \textit{original}, consisted in using images 'as is', i.e. without any pre-processing. The second option, namely \textit{cropped}, consisted in a simple image crop so that image borders were removed. The third option, namely \textit{ReSampFid}, consisted in a resampling of the scanned images based on the four fiducial marks of the corners so that the resulting images shared the same geometry, with fiducial marks centres constituting the corners of the resampled images. The third option was done with the ReSampFid tool of the MicMac open source photogrammetric suite \citep{Rupnik2017} and an ad-hoc tool for automated fiducial marks detection developed with ImageJ.

\subsubsection{Time-SIFT}

The proposed Time-SIFT method relies on the properties of the SIFT-like algorithms used in SfM workflows. These algorithms compute keypoints that are invariant to image rotation and scale and hence robust across a substantial range of affine distortions, addition of noise and change in illumination \citep{Lowe2004, PhotoscanAlgo}. 

By making the assumption that a sufficient number of keypoints remain invariant across each time period, the proposed Time-SIFT method brings a novelty in the photogrammetric processing of multi-temporal datasets. It consisted in processing, at the very first step, all images in a single block for the estimation of interior and exterior orientations. Images of different dates were then separated and the dense matching steps were done with images of the same date.

\subsubsection{Use of additional manually marked points}\label{sec:additionnal_points}

Different numbers and types of additional - relatively to the automatic tie points - manually marked points were used in the different processing scenarios, with a minimal set of 9 GCPs in order to scale and orient the estimated DEMs and DoDs.

The first option, namely \textit{9 GCPs}, corresponded to this minimal set of 9 additional manual GCPs. For the second option, namely \textit{200 tie points}, we added manually marked tie points to reach a total of 200 points.  For the third option, namely \textit{200 GCPs}, ground coordinates of all the 200 points were also given and used in the SfM workflow. For all these three options, the manually marked points were added to the set of automatic tie points.

The additional 200 points consisted in carefully chosen permanent ground features, for instance road intersections or little rocks, which were manually pointed in the images. Corners of buildings or big rocks were avoided so that the less possible variation of the $z$ coordinate would occur around the chosen points. At first a regular sampling of 200 locations over the whole image block area was determined and each of the 200 points was chosen within this regular grid to ensure an homogeneous point density (Figure \ref{fig:test_site_DEMs_GCPs_transects}). Each of the 9 or 200 points was manually pointed in the aerial images of the four dates. When used as GCP, $(x,y,z)$ ground coordinates were necessary. They were collected on the french geographic survey portal on which orthorectified imagery and DEM can be browsed (\url{https://www.geoportail.gouv.fr/}). Empirical observations during the marking of the 200 points in multi-date aerial images resulted in estimated accuracy values of 20 m for ground accuracy and 5 pixels for image accuracy (Table \ref{tab:photoscanfixed} above). The total proceeding time for pointing and assigning coordinates on 200 locations in 163 images was about twenty hours.

\subsubsection{Autocalibration}

The autocalibration method aimed at taking into account - or not - the fact that all images of a single date were taken by the same camera. The option \textit{by date} consisted in using a single camera model for all the images of a given date whereas the second option \textit{by image} consisted in using one camera model by image. In all cases, we kept the parameters suggested by PhotoScan Pro\textsuperscript{\copyright} for lens distortion estimation (Table \ref{tab:photoscanfixed}).

\subsection{Quality assessment}

\subsubsection{Validation data}

Two external DEMs obtained at different dates and with different methods were used for quantitative independent validation. A photogrammetric survey of the area realised in 2012 by Topogeodis provided a DEM with a pixel resolution of 5 m (the larger DEM on Figure \ref{fig:test_site_DEMs_GCPs_transects}). Its planimetric accuracy is estimated at 50 cm. The altimetric accuracy is estimated at 30 cm in urban zones and at 1 m in mountainous and woodlands areas.
In 2001, a lidar survey was conduced by Geolas Consulting on a subaera of 7 km\textsuperscript{2} (the smaller area bordered by dashed lines in Figure \ref{fig:test_site_DEMs_GCPs_transects}). From these data, a DEM with a pixel resolution of 1 m and an estimated altimetric accuracy of 30 cm was derived.

\subsubsection{Quantitative indices}

The impact of the different processing options on the results was evaluated with four different metrics. The first metric, 'Image block RMSE', was based on the internal coherence of the image block estimated geometry. It consisted in calculating the global RMSE of the bundle block adjustment. It was computed for the 144 image blocks. The second metric, 'Coarse DEM error', was computed using as reference the 2012 photogrammetric DEM, which covers the whole test site (Figure \ref{fig:test_site_DEMs_GCPs_transects}). The reference DEM and all the 144 SfM-MVS DEMs were first resampled at 50 m with bilinear interpolation. The 'Coarse DEM error' was then computed as the mean absolute difference between simulated and reference DEMs. This metric was chosen to evaluate whether the computed DEMs of each date were coherent with the actual topography. This metric was computed at 50 m resolution considering that potential 3-D changes would be negligible at this resolution and has hence been computed for each date. The third metric, 'Fine DEM error' was computed with the 2001 lidar DEM, which covers a smaller part of the test site as shown on Figure \ref{fig:test_site_DEMs_GCPs_transects}. It aimed at checking whether the fine scale topography was correctly estimated by the SfM-MVS DEMs. As for the 'Coarse DEM error', the metric was computed as the mean absolute difference between simulated and reference DEMs. Considering that, at this scale, 3-D change could not be neglected, this metric was only evaluated on the 36 DEMs computed with the images of the year 2001. The fourth metric, 'DoD quality', corresponded to the internal quality of the computed DoDs, estimated by the mean absolute value of the DoD. These metrics are summarised in Table \ref{tab:quantitative_indices_descr}.

\begin{table}
\centering
\scriptsize{
\begin{tabular}{l c c p{2.7cm}}
Name & Unit & Samples & Description \\
\hline
Image block RMSE & pixels & 144 Image blocks & Bundle block adjustment RMSE\\
Coarse DEM error & meters & 144 DEMs & Coarse resolution mean absolute difference between the resampled SfM-MVS DEMs and reference DEM\\
Fine DEM error & meters & 36 DEMs & Mean absolute difference between the 2001 SfM-MVS DEM and the fine resolution reference lidar DEM \\
DOD quality & meters & 108 DoDs & Mean of the DoD absolute values\\
\end{tabular}
}
\caption{Metrics used for the evaluation of the different processing options.}
\label{tab:quantitative_indices_descr}
\end{table}

For all these metrics, we examined whether the different processing options described in section \ref{sec:processing_options} had a significant impact on the distribution of the metric or not. Due to strong skew of the metrics distributions, we used the Kruskal-Wallis test to determine the significance of the observed differences.

\subsubsection{Qualitative analyses of the DoDs}

We visually examined all 108 DoDs at coarse scale to explain the quantitative results obtained with scalar metrics. We also examined topographic profiles of the different DEMs to better understand the results obtained on the DoDs. Theses profiles were extracted on the two transects drawn on Figure \ref{fig:test_site_DEMs_GCPs_transects}.
The best DoDs - in the sense of the 'DoD quality' metric as described above - obtained with the less external information (only 9 GCPs) were finally visually examined at full resolution. The detected change signals were associated to actual 3-D changes that could be confirmed by other evidences, either external documentation, either image information. Amongst the 3-D changes detected, we examined 3-D changes linked to anthropogenic activity, such as urban growth, civil engineering around highway, quarries and backfills. These analyses aimed at validating the potential of the Time-SIFT method for detecting and characterising 3-D changes with frugal ground information.

\section{Results}

\subsection{Quantitative assessment}

\begin{table}
\Rotatebox{90}{%
\scriptsize{
\begin{tabular}{ll ccl ccl ccl ccl}
& & \multicolumn{3}{c}{Image  block  RMSE (pix)} & \multicolumn{3}{c}{Coarse  DEM error (m)}  & \multicolumn{3}{c}{Fine  DEM  error (m)} & \multicolumn{3}{c}{DoD  quality (m)}\\
\multicolumn{2}{c}{Processing options} & \multicolumn{3}{c}{samples : 144} & \multicolumn{3}{c}{samples : 144} & \multicolumn{3}{c}{samples : 36 (2001 images)} & \multicolumn{3}{c}{samples : 108}\\
& & median & IQR & p-value  & median & IQR & p-value &  median & IQR & p-value & median & IQR & p-value\\
\hline
\multirow{3}{1.5cm}{Image pre-processing}
& \textit{original} & 0.69 & 0.50 & & 12.5 & 13.5  & & 10.7 & 3.1 & & 8.4 & 28.2 & \\
& \textit{cropped} & 0.76 & 0.29 & 0.036\textsuperscript{*} & 23.6 & 55.1 & 4.7e-04\textsuperscript{***} & 39.1 & 67.4 & 5.9e-04\textsuperscript{***} & 14.2 & 25.1 & 0.20 \\
& \textit{ReSampFid} & 0.63 & 0.24 &  & 5.7 & 10.2 &  & 2.5 & 3.3 & & 7.4 & 24.3 & \\
\hline
\multirow{2}{*}{Time-SIFT}
&\textit{without} & 0.76 & 1.05 & \multirow{2}{*}{2.6e-03\textsuperscript{**}} & 20.4 & 29.9 & \multirow{2}{*}{5.0e-08\textsuperscript{***}} & 11.2 & 33.4 & \multirow{2}{*}{0.27} & 25.0 & 65.0 & \multirow{2}{*}{7.8e-13\textsuperscript{***}} \\
&\textit{with} & 0.64 & 0.23 & & 7.5 & 5.2 &  & 6.2 & 5.4 & & 2.1 & 5.5 & \\
\hline
\multirow{3}{2cm}{Additional manually marked points}
&\textit{9 GCPs} & 0.68 & 0.24 & & 9.1 & 18.1 & & 8.8 & 12.8 & & 10.1 & 23.3 & \\
&\textit{200 tie points} & 0.67 & 0.26 &  0.062 & 9.7 & 16.6 & 0.82 & 9.8 & 25.7 & 0.62 & 11.0 & 22.8 & 0.50\\
&\textit{200 GCPs} & 0.74 & 0.53 & & 12.3 & 64.1 & & 6.8 & 27.4 & & 14.4 & 57.3 & \\
\hline
\multirow{2}{*}{Autocalibration}
& \textit{by image} & 0.69 & 0.27 & \multirow{2}{*}{0.83} & 11.8 & 24.9 & \multirow{2}{*}{0.31} & 8.5 & 36.5 & \multirow{2}{*}{0.90} & 12.0 & 16.8 & \multirow{2}{*}{0.49} \\
& \textit{by date} & 0.70 & 0.30 & & 8.9 & 26.4 & & 9.7 & 12.0 & & 14.3 & 39.7 & \\
\end{tabular}
}
}
\caption{Effect of the processing options on the image block quality, DEM quality and DoD quality. For each metric, the total sample number is recalled and the median value, inter-quantile range (IQR) and the p-value of the Kruskal-Wallis test for the significance of the difference of medians is given.}
\label{tab:quantitative_indices_res}
\end{table}

The main parameters of the metrics distributions are shown in Table \ref{tab:quantitative_indices_res}. For more details, distribution boxplots are given in Appendix (Figure \ref{fig:boxplots_plus}). These results exhibit two groups of processing options that had a significant impact on all metrics: the image pre-processing options and the proposed Time-SIFT method. The most significant result is that using the Time-SIFT method led to a median DoD quality value of 2.1 m, within an inter-quartile range of 5.5 m, compared to a median DoD quality value of 25.0 m (inter-quartile range of 65.0 m) for all the 54 DoDs computed without the Time-SIFT method. For all metrics, using the Time-SIFT method allowed for achieving a better quality, this result being significant for 3 metrics over 4. Second, using the \textit{ReSampFid} option for image pre-processing also significantly and positively impacted the results by lowering the Coarse DEM error and Fine DEM error to 5.7 m and 2.5 m, respectively. Third, and as testified by the negligible significances for this group of options, adding manually marked points had a negligible impact on the quality of the DEMs and of the DoDs. Furthermore, results obtained with these different options were contradictory between metrics. With even worse p-values, autocalibration options also seemed to have had a negligible impact on all metrics.

In absolute values, using the Time-SIFT method is also the processing option which resulted in the best DoD quality, both in terms of median and inter-quartile range. In addition, the \textit{ReSampFid} option resulted in the best individual DEMs both at fine and coarse scale. Our experiments hence showed that the Time-SIFT method, and secondly the image pre-processing, had the most significant impact on the DEMs and the DoDs. The \textit{ReSampFid} option seems most important to obtain best individual DEMs whereas best DoDs were obtained by using the Time-SIFT method. This may be due to the fact that accurate DEMs do not necessarily resulted in accurate DoDs when DEMs geometries were not coherent enough.


\subsection{Qualitative assessment}

\begin{figure*}
	\centering
	\includegraphics[width=\linewidth{}]{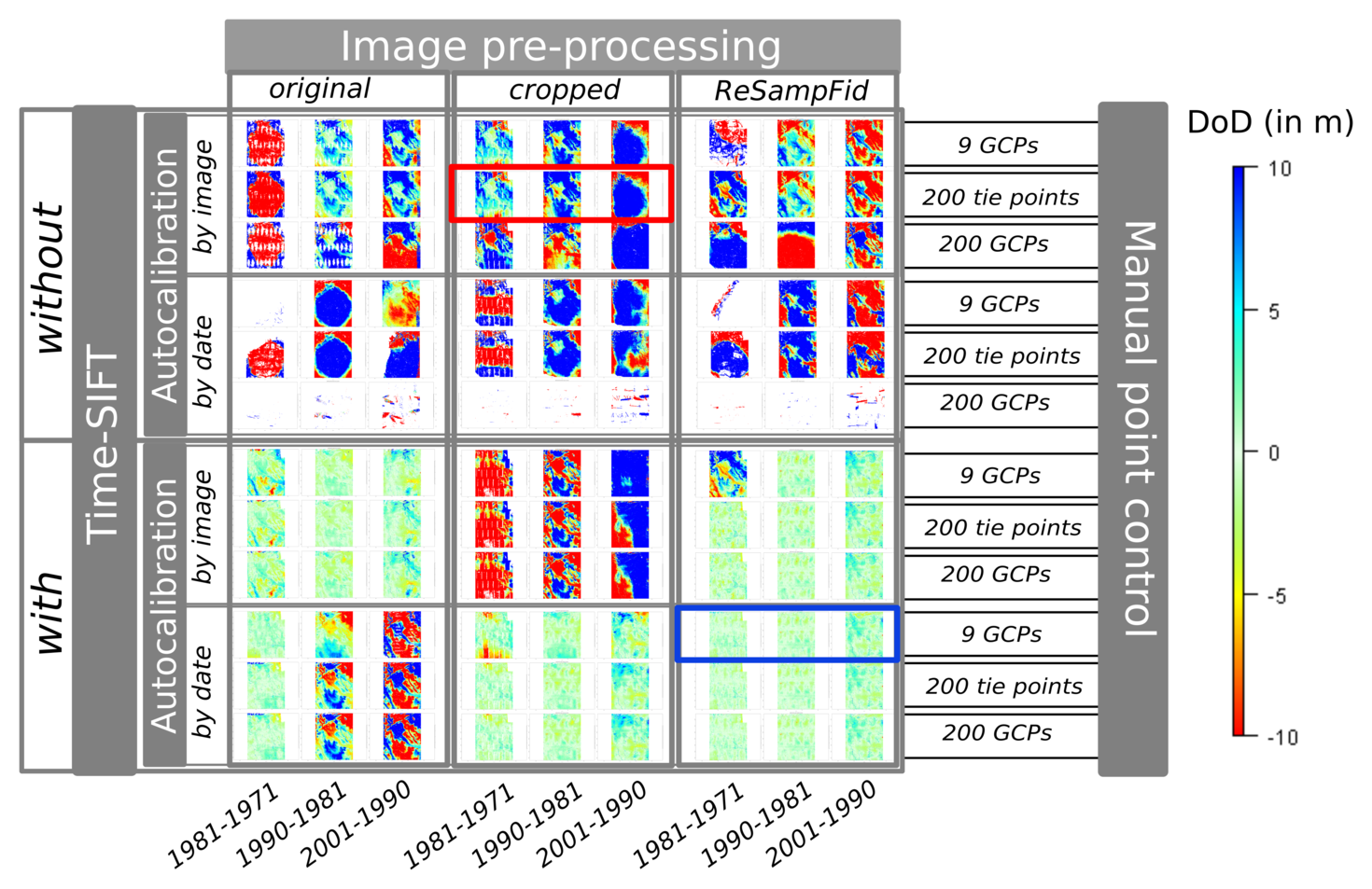}
	\caption{DoDs between consecutive periods for each combination of the processing options. All DoDs are represented with the same colour scale truncated at $\pm$10 m. Red and blue rectangles framed respectively the best model without using the Time-SIFT method, and the best model with the Time-SIFT method.}
    \label{fig:DoDs}
\end{figure*}

Figure \ref{fig:DoDs} shows the thumbnails of all the 108 DoDs obtained with the different processing options. First, a few DoDs show spatial patterns that are very far from what could be expected. These flawed DoDs correspond to situations where photogrammetric processing converged to an erroneous solution, most likely since the SfM processing step. For some combinations of processing options, such as, for instance, using \textit{200 GCPs}, not using the Time-SIFT method and doing autocalibration \textit{by date}, almost no solution could be found, regardless of the image pre-processing options.

A strong concave or convex 'doming' effect \citep{JamesRobson2014} can be observed for several DoDs that where computed without the Time-SIFT method. These findings give indications on the problems that may have occurred during the SfM processing step, in particular erroneous autocalibration.\\

\begin{figure*}
	\centering
	\includegraphics[width=\linewidth{}]{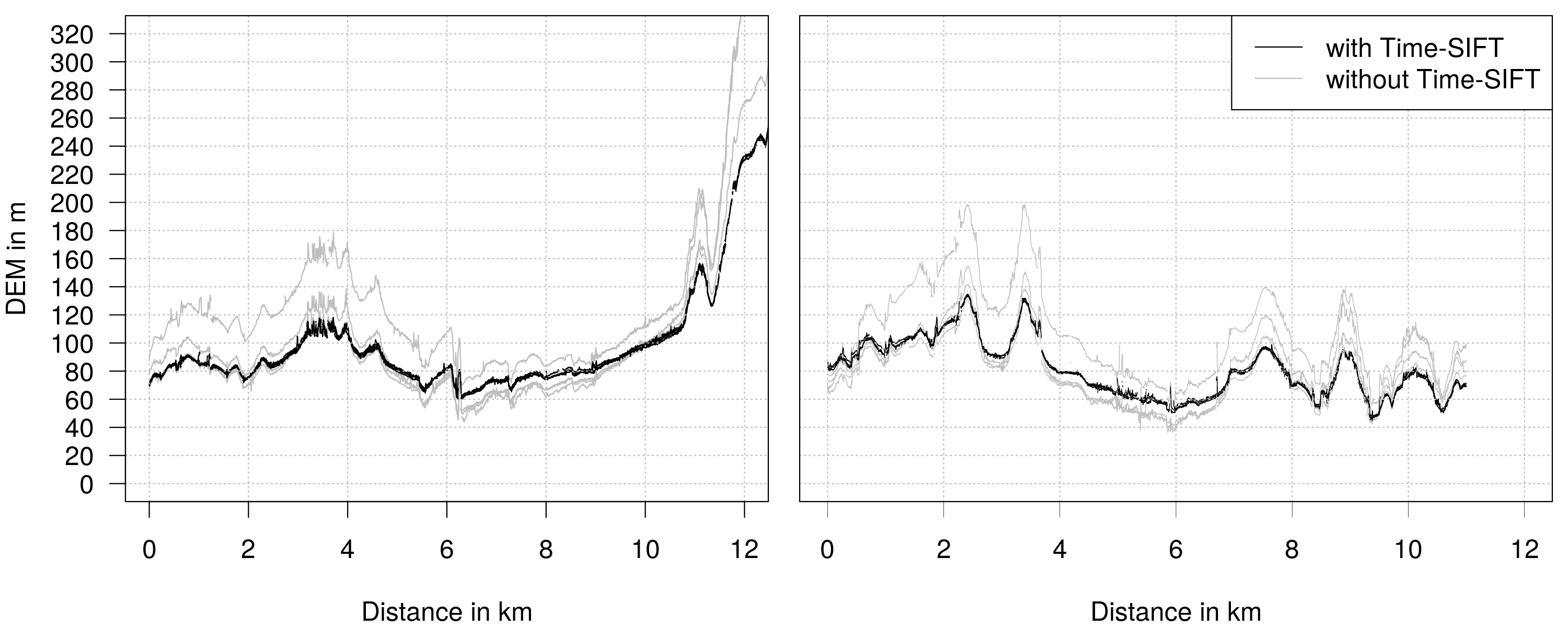}
	\caption{Profiles comparison between the bests DEMs computed respectively with (in black) and without (in grey) the Time-SIFT method. Each line corresponds to a different date. Left, the East-West transect ; right, the North-South transect  (see transects on Figure \ref{fig:test_site_DEMs_GCPs_transects}).}
    \label{fig:transects}
\end{figure*}

Figure \ref{fig:transects} compares - on two transects - topographic profiles of the best DEMs computed on the one hand \textit{with} the Time-SIFT method and on the other hand \textit{without} the Time-SIFT method. The DoDs corresponding to these DEMs are marked respectively by blue and red rectangles in Figure \ref{fig:DoDs}. Comparison of theses profiles shows that the topographic differences between DEMs computed with different options exhibit low-frequency trends rather than local differences. Indeed, topographic shapes are locally correctly rendered within all DEMs, but with large and non-stationary biases in the DEMs computed without the Time-SIFT method. \\

Finally we selected the best DoDs among those obtained by using the Time-SIFT method and the less additional external information. It corresponded to the association of \textit{ReSampFid}, \textit{9 GCPs} options and autocalibration \textit{by date}. On these DoDs at full resolution, we sought 3-D changes signals and inspected them.

\begin{figure}
\centering
    \includegraphics[width=\linewidth]{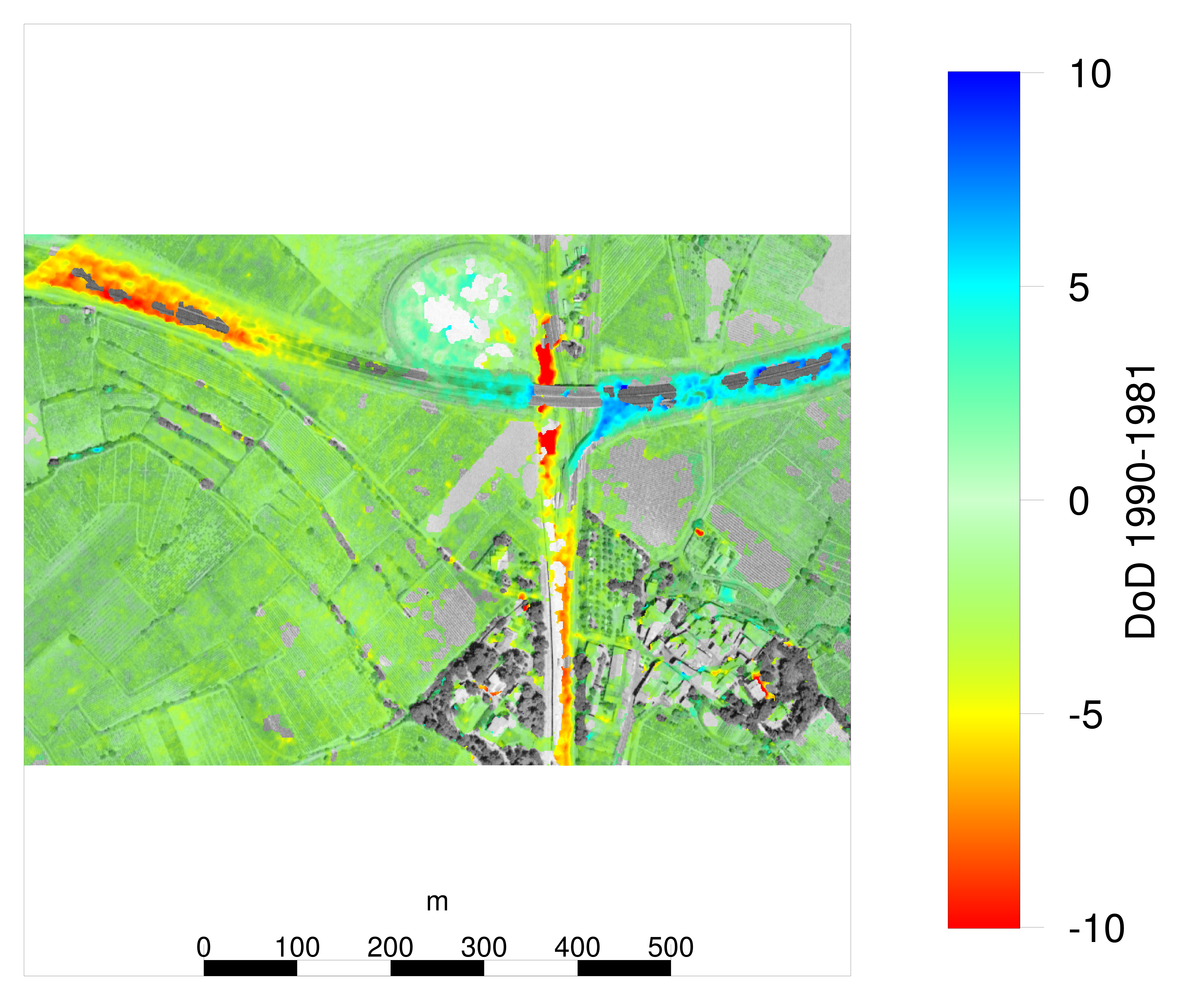}
  \caption{3-D changes associated to highway construction. The 1990-1981 DoD is superimposed on the 1990 orthophoto. Colourless areas correspond to areas where no value was computed in the DoD. The DoD was estimated with the Time-SIFT method, using \textit{9 GCPs}, autocalibration with a single camera \textit{by date} and the \textit{ReSampFid} pre-processing.}
  \label{fig:highway}
\end{figure}

First, we looked for evidences which would demonstrate unambiguously that these 3-D changes are not processing artefacts. Figure \ref{fig:highway} shows excavation (red) and fill (blue) patterns that are clearly superimposed on the newly constructed highway. This demonstrates the capability of the detecting 3-D changes in DoDs computed with the Time-SIFT method and limited external information. This figure also shows that, even if most of the area contains 3-D difference information, processing may fail. In this zone, problems were faced in low-texture areas, on vegetation, on some parts of urban areas, and on some field plots.

\begin{figure}
\centering
  \includegraphics[width=\linewidth]{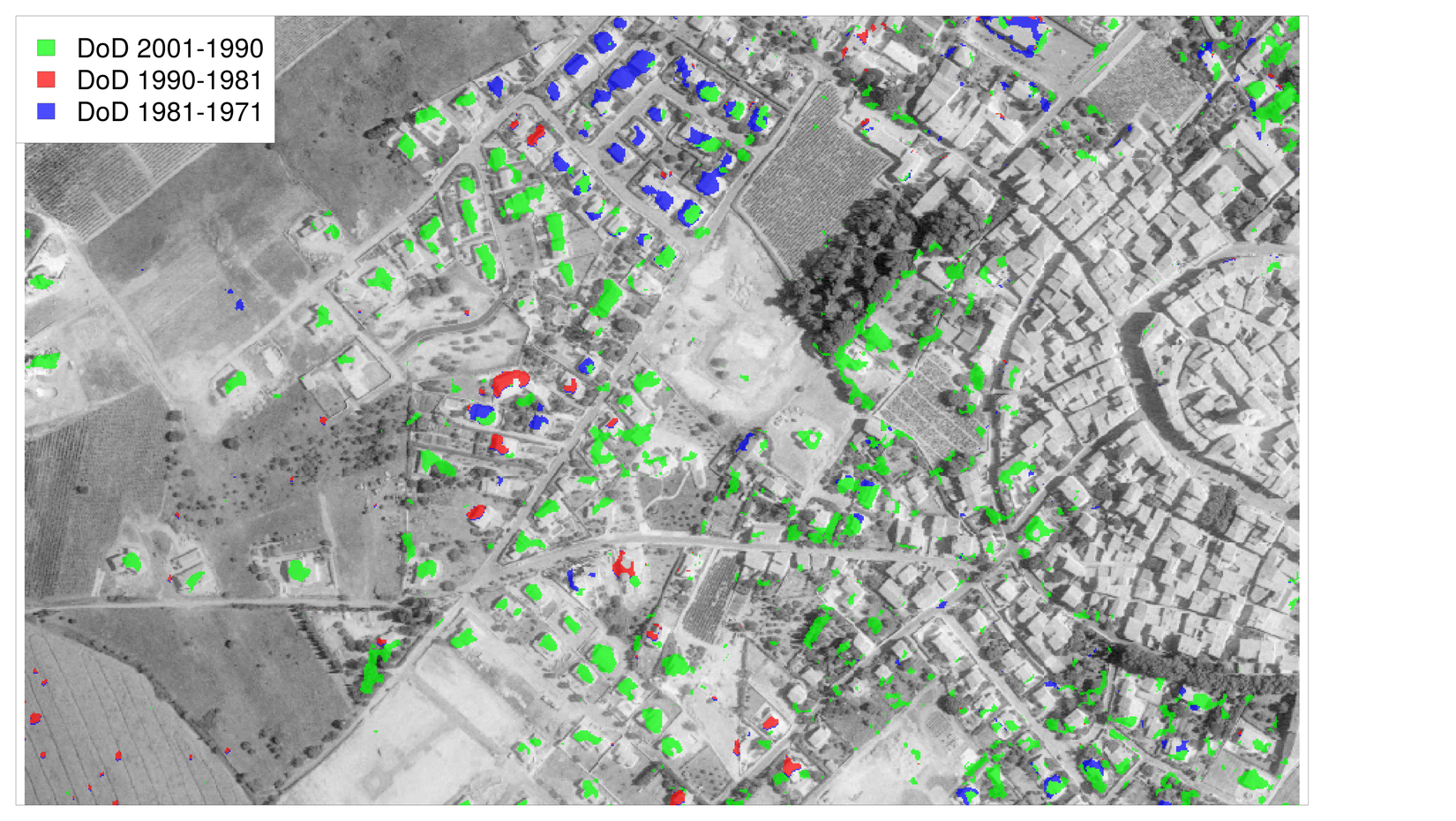}
  \caption{Superimposition of thresholded DoDs (values greater than 2.5 m) onto the 2001 orthophoto (gray levels). The DoDs were estimated with the Time-SIFT method, using \textit{9 GCPs}, autocalibration with a single camera \textit{by date} and the \textit{ReSampFid} pre-processing. \textit{with} Time-SIFT.}
  \label{fig:urban}
\end{figure}

Figure \ref{fig:urban} allows to detail the potential of these DoDs in terms of spatial resolution. It shows the DoDs thresholded above 2.5 m and superimposed on the 2001 orthophoto. Rectangular patterns in these thresholded DoDs show strong correlations with the floorspace of the houses that were built within the period. This figure also shows artefacts which may be either small and isolated patches, either thin patterns at the edges of sharp objects.

\begin{figure}
\centering
  \includegraphics[width=8cm]{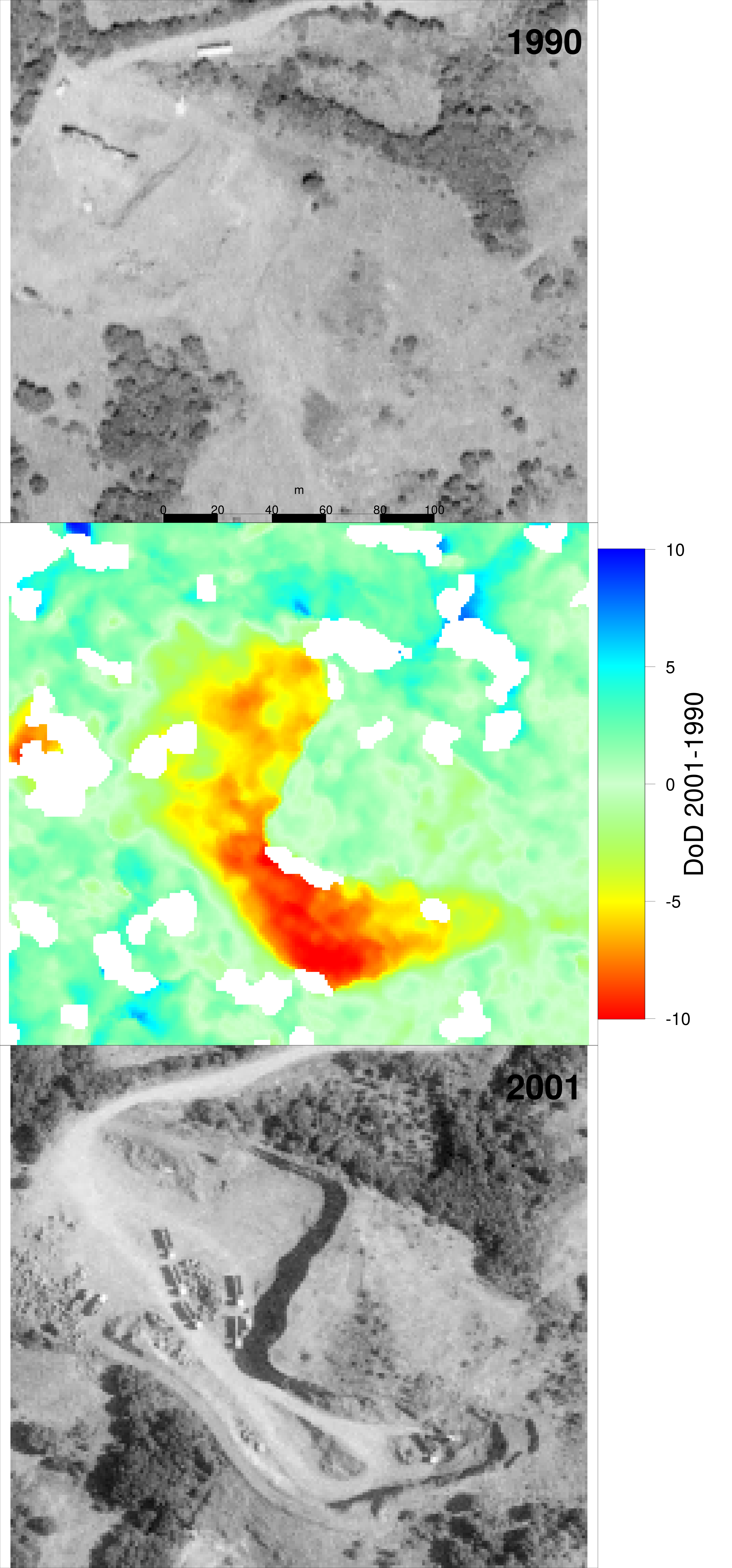}
  \caption{3-D changes associated to excavation activities in a quarry. Top : the 1990 orthophoto. Middle, the 2001-1990 DoD. Colourless areas correspond to areas where no value was computed in the DoD. The DoD was estimated with the Time-SIFT method, using \textit{9 GCPs}, autocalibration with a single camera \textit{by date} and the \textit{ReSampFid} pre-processing. Bottom : the 2001 orthophoto}
  \label{fig:quarry}
\end{figure}

\begin{figure}
\centering
  \includegraphics[width=8cm]{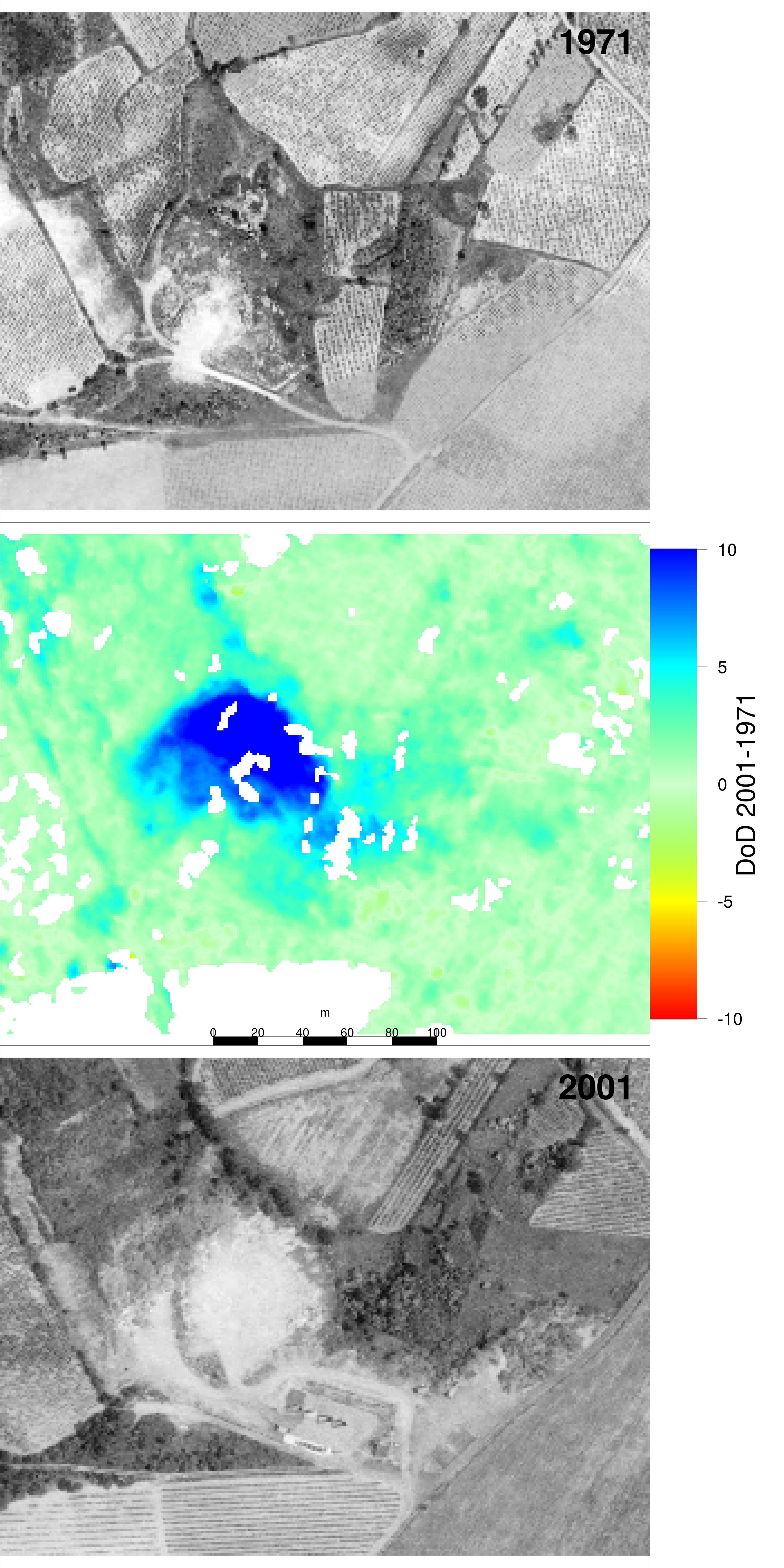}
  \caption{3-D changes associated to ground filling. Top : the 1971 orthophoto. Middle, the cumulated 2001-1971 DoD. Colourless areas correspond to areas where no value was computed in the DoD. The DoD was estimated with the Time-SIFT method, using \textit{9 GCPs}, autocalibration with a single camera \textit{by date} and the \textit{ReSampFid} pre-processing. Bottom : the 2001 orthophoto}
  \label{fig:backfill}
\end{figure}

Figures \ref{fig:quarry} and \ref{fig:backfill} allows to detail the potential of the obtained DoDs in terms of volume quantification. These figures represent respectively the ground excavating and filling of a limited area. These two 3-D changes couldn't be detected from orthophotos alone, in terms of volume quantification and even for the delineation of the impacted zone. This is especially depicted in Figure \ref{fig:backfill}, where changes in texture have a larger extent than changes in surface height.

\section{Discussion}

The objective of this work was to determine the interest of a new method, called Time-SIFT, to compute informative DEMs of difference with limited photogrammetric expertise and almost exclusively image information.\\

After a thorough exploration of different processing strategies, we showed that the use of the Time-SIFT method allowed for computing DoDs in which actual 3-D changes could be detected and quantified. Moreover, the Time-SIFT method played the most significant role to ensure the computation of high-quality DoDs. To the best of our knowledge, automated computation of multi-temporal feature points for DoD estimation by feeding SIFT-like algorithms with multi-temporal datasets was never assessed before. \cite{Hapke2005}, \cite{Korpela2006} and \cite{Dewitte2008} already used multi-temporal tie points for this purpose but these points were manually marked. \cite{Fox2008} computed tie points between images of different dates and even thought about a method similar to the one proposed in this paper in what they called a simultaneous 'grand adjustment'. However, and even if these authors appreciated this possible method as elegant, they rejected this option in order to keep the quality of the reference block geometry and to avoid difficult posterior error checks. Thus, and even if temporal SIFT methods are common in computer vision for the processing of videos, to the best of our knowledge only two authors had proposed studies with experiments showing some similarity with our method. In a study aiming at characterising the yearly evolution of the snowline, \cite{Vargo2017} used SfM photogrammetry to process a multi-temporal dataset of aerial imagery with at least one image a year from 1981 to 2017. Due to the fact that for some years only one image was available, they have had to put images of different dates in the same block. Nevertheless, they also have had to mask snow and ice out of all images so that the algorithm could estimate the orientations of these isolated images. \cite{Chanut2017} also successfully used a SfM algorithm to compute tie points between images of different dates in a study that aimed at quantifying 3-D movements on a landslide. Multi-temporal tie points were used to initialise a dense correlation step. Mass movements were finally quantified from the results of the dense correlation.

The second most important finding is that pre-processing of scanned analogue imagery also plays a significant role in the processing steps of SfM-MVS. This is coherent with the existing literature that dealt with the use of archival aerial imagery with SfM algorithms. \cite{Salach2017} developed a specific software to resample scanned analogue photographs relatively to information given by fiducial marks. These authors tested different processing strategies : with or without pre-processing, and different GCPs number and configurations. They reported that initial estimates of image orientations with this pre-processing were better than those obtained without this pre-processing. More, results with pre-processing and a minimal GCP set were similar to the results without pre-processing but with a complete set of GCPs. The latter result is hence corroborated by the results of our study. \cite{Goncalves2016} used a manual affine transformation computed on the position of the fiducial marks to register the scanned analogue images between each other. This allowed the use of archival aerial imagery in SfM workflow for automated mosaicing and DEM production. The RMSE of the obtained DEM, of 4.7 m, is of the same order of magnitude than the DEM errors estimated in our study. \cite{Nocerino2012} also exploited the fiducial marks to settle an image reference coordinate system at the geometrical center of the fiducial marks. RMSE estimated on the $z$ coordinates of check points ranged from 7 m to 10 m. Our results compares favourably with this estimated decametric accuracy.

The third finding can be seen as more unexpected : our experiments showed that the use of a relatively high number of  added manual GCPs and/or tie points seems to play a negligible role on the image blocks, DEMs, and DoDs quality. This finding seems to be in contradiction with the results of previous studies based on the thorough use of GCPs to compute DoDs from archival aerial imagery. Indeed, \cite{Hapke2005,Zanutta2006,Dewitte2008,Fox2008,Micheletti2015,Papworth2016,Mertes2017} successfully used GCPs of different sources to ensure a geometrical consistence between DEMs before subtracting them to each other. This apparent contradiction may be explained by the fact that these studies used a method that significantly differs from ours. Contrarily to these studies, we did a fully automatic photogrammetric processing, from image block orientation to DEM extraction and DoD computing. With a view to be able to apply our method on large image data sets, we didn't manually check the results at any intermediary steps. As a consequence we eventually missed gross GCP errors such as the one mentioned for instance by \cite{Verhoeven2016}, who iteratively eliminated the GCPs that caused largest errors. In our case, this uncontrolled use of additional manually marked points resulted in flawed DoDs for some processing options combinations. The synthesis of our results and the ones of the literature is that both cumbersome manual marking in all images and a time-consuming manual check of image orientation results are required so that manually marked GCPs would improve the photogrammetric processing.\\

To our opinion, the fact that the Time-SIFT method allows for a correct computation of DoDs has at least two explanations. The first one is linked to the use of tie points that were computed between images of different dates. Image block geometries of different dates are hence coherent by construction and consequently, DEMs spatial references are consistent and allow for a proper estimation of their difference. The second is linked with autocalibration. We postulate that the Time-SIFT method has also benefits for the SfM autocalibration step. Autocalibration of archival aerial imagery may indeed be sensitive, as noted for instance by \cite{Aguilar2013}, and can be even more difficult with SfM algorithms. This is due to the geometry of the archival images acquisitions and its inadequacy with the requirements of SfM algorithms. First, overlaps in archival aerial imagery are small (60\%) compared to the overlaps required for SfM algorithms (80\%, or even 90\%). Second, in our study, with altitudes higher than 2700 meters and height difference on ground lesser than 400 m, ground is mostly flat relatively to flying height. As a consequence, acquisition geometry of archival aerial imagery constitutes a non-ideal case for autocalibration, with strong possible correlations between estimated interior and exterior orientations, in particular flying height and focal length. This correlation may result in the doming effect observed by \cite{JamesRobson2014} on UAV images and that we also observed in some processing scenarios. Our hypothesis is that autocalibration in a same image block of the four cameras used at four different dates, with automatically computed multi-temporal tie points, allows for a lesser correlation of interior and exterior orientation and hence a better result.\\

Qualitative analyses of the DoDs obtained with the Time-SIFT method also demonstrated its potential. DoDs obtained using a limited number of GCPs indeed unambiguously detected and quantified past 3-D changes. As orthophotos greatly changed in texture and gray levels from year to year due to the significant evolution of land management and land cover on the area \citep{Vinatier2018}, the use of DoDs allowed to detect 3-D changes that would probably be missed by orthophoto analysis. It is also worth noting that the magnitudes of the changes mapped by our method are small relatively to the changes observed with DoDs on volcanoes \citep[e.g.][]{Gomez2014} and on glaciers \citep[e.g.][]{Mertes2017,Molg2017}. To the best of our knowledge, only \cite{Bakker2017} presented results with 3-D changes smaller than $\pm$3 meters. To achieve this precision, they used an external lidar dataset and the a priori knowledge of stable areas for the fine co-registration of multi-temporal DEMs. Furthermore, the study of anthropogenic changes from DoDs rather than orthophotos is rarely encountered in the literature, except in \cite{Sevara2017}, who also succeeded in mapping 3-D changes related to quarries thanks to a fine co-registration of each DEM with an external lidar DEM prior to photogrammetric DEMs differencing.\\

With an extensive test of different processing scenarios, we showed that the Time-SIFT method may be advantageous for a first blind exploration of archival aerial imagery for detection of 3-D changes. Even if our studies already covers a time span of three decades and an area of 170 km\textsuperscript{2}, the results may be different with other time spans and/or on other sites, in particular on sites that would exhibit even bigger changes than ours. Anyhow, the Time-SIFT method can conveniently and inexpensively be used at first, with no other information that image information, in order to detected zones with 3-D changes. This approach - using the Time-SIFT method without any GCP - was already successfully tested \citep{Feurer2017SFPT}. It can favourably be used with datasets of thousands of images. Users interested in a more detailed and accurate description of 3-D changes may then use additional information and/or a dedicated workflow focused on the zones where 3-D changes would have been detected with the Time-SIFT method.

Other future works should test the robustness of the Time-SIFT method to bigger time spans and to different landscapes types. Some improvements may also be done on the interior orientation step. Indeed, by using more fiducial marks than the 4 marks of the corners, a better model of camera inner orientation may be estimated. Finally, there is still a need to find methods that would allow for the use of archival aerial imagery scanned with desktop scanners instead of photogrammetric scanner. This would necessitate to use more information from the scanned analogue image borders for instance, in order to determine a model of scanner deformation.

\section{Conclusion}

In this paper we proposed and tested a new method, termed Time-SIFT, which allows for the detection of 3-D changes from archival aerial imagery with almost exclusively image information. We showed that the Time-SIFT method allowed for the computing of informative DEMs of difference, in which 3-D changes could be detected and characterised. Due to the fact that the Time-SIFT method can be applied automatically and does not require neither photogrammetric expertise nor additional data, this new method may pave the way to a more extensive use of the worldwide aerial imagery archive. This method hence constitutes an additional tool in geosciences for the detection of past 3-D changes. This work set the stage of two types of new studies. First this method has now to be tested in different contexts and with other images datasets in order to determine its robustness. Second the worldwide archival aerial imagery can be explored with the Time-SIFT method so that past 3-D changes may be discovered or characterised.

The base hypothesis of our work was to lessen as much as possible the use of external data and to make available to the broader audience the multi-temporal 3-D information of archival aerial imagery. Our work hence focused on easily available and/or inexpensive data and methods. It is indeed a prime criteria so that it can further be applied on larger datasets within as much different contexts as possible and by the widest community.\\

\clearpage

\section*{References}
\bibliographystyle{elsarticle-harv}
\bibliography{Biblio_papier_TimeSIFT.bib}

\appendix
\section{Distribution of error metrics}
\begin{landscape}
\begin{figure*}
	\centering
	\includegraphics[width=\linewidth{}]{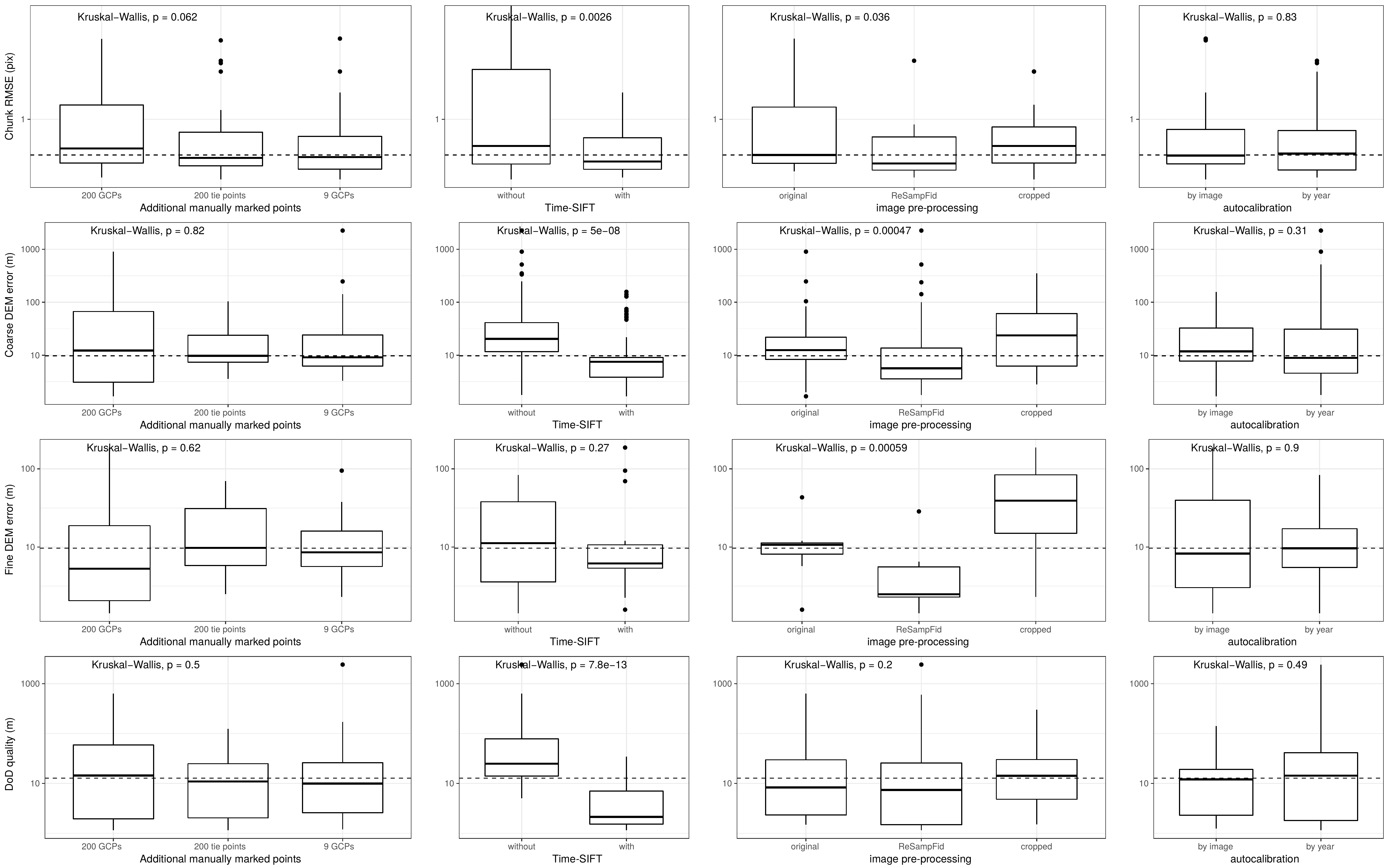}
\caption{Boxplots of (a) alignment quality, (b,c) the difference between computed and reference DEM, respectively the coarse globale and fine local lidar DEM and (d) DoD quality . P-values issued from Kruskal-Wallis tests comparing each modality side-by-side are given.}
    \label{fig:boxplots_plus}
\end{figure*}
\end{landscape}

\end{document}